\documentclass[10pt,twocolumn,letterpaper]{article}

\usepackage{cvpr}

\usepackage[dvipsnames]{xcolor}

\usepackage{graphicx}
\usepackage{amsmath}
\usepackage{amssymb}
\usepackage{booktabs}
\usepackage{algorithmic}
\usepackage{algorithm}
\usepackage{multirow}
\usepackage[pagebackref,breaklinks,colorlinks,citecolor=cvprblue]{hyperref}

\definecolor{cvprblue}{rgb}{0.21,0.49,0.74}
\def\confName{WACV}
\def\confYear{2025}

\title{ERUP-YOLO: Enhancing Object Detection Robustness for Adverse Weather Condition by Unified Image-Adaptive Processing}

\author{Yuka Ogino, Yuho Shoji, Takahiro Toizumi, Atsushi Ito\\
NEC Corporation\\
{\tt\small yogino@nec.com, yuho-shoji@nec.com, t-toizumi\_ct@nec.com, ito-atsushi@nec.com}
}
\begin{document}
\maketitle
\begin{abstract}
We propose an image-adaptive object detection method for adverse weather conditions such as fog and low-light. Our framework employs differentiable preprocessing filters to perform image enhancement suitable for later-stage object detections. Our framework introduces two differentiable filters: a Bézier curve-based pixel-wise (BPW) filter and a kernel-based local (KBL) filter. These filters unify the functions of classical image processing filters and improve performance of object detection. We also propose a domain-agnostic data augmentation strategy using the BPW filter. Our method does not require data-specific customization of the filter combinations, parameter ranges, and data augmentation. We evaluate our proposed approach, called Enhanced Robustness by Unified Image Processing (ERUP)-YOLO, by applying it to the YOLOv3 detector. Experiments on adverse weather datasets demonstrate that our proposed filters match or exceed the expressiveness of conventional methods and our ERUP-YOLO achieved superior performance in a wide range of adverse weather conditions, including fog and low-light conditions.
\end{abstract}


\section{Introduction}
\begin{figure}[t]
    \centering
    \begin{tabular}{cc}
    \includegraphics[width=0.47\hsize]{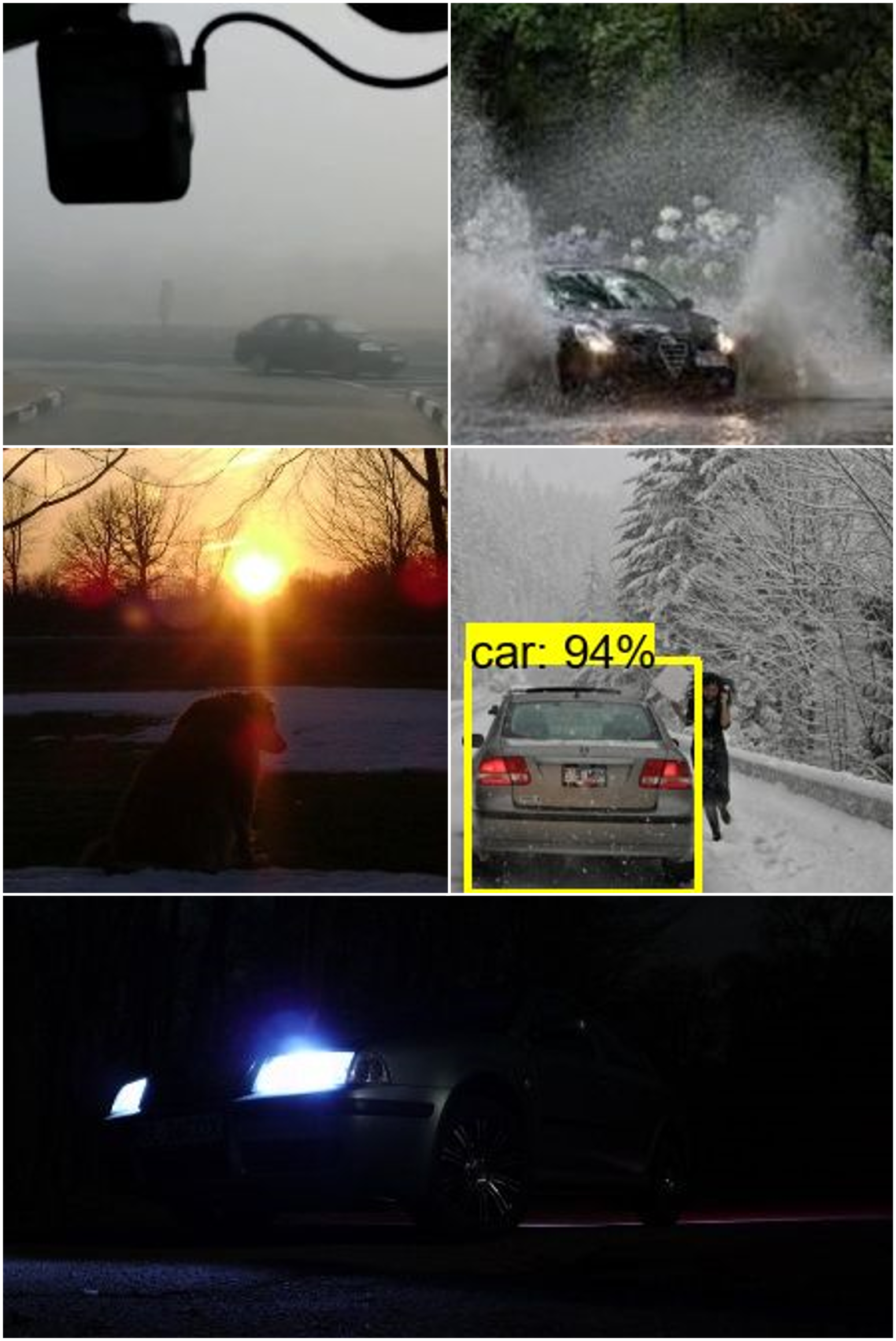}&\hspace{-13pt}
    \includegraphics[width=0.47\hsize]{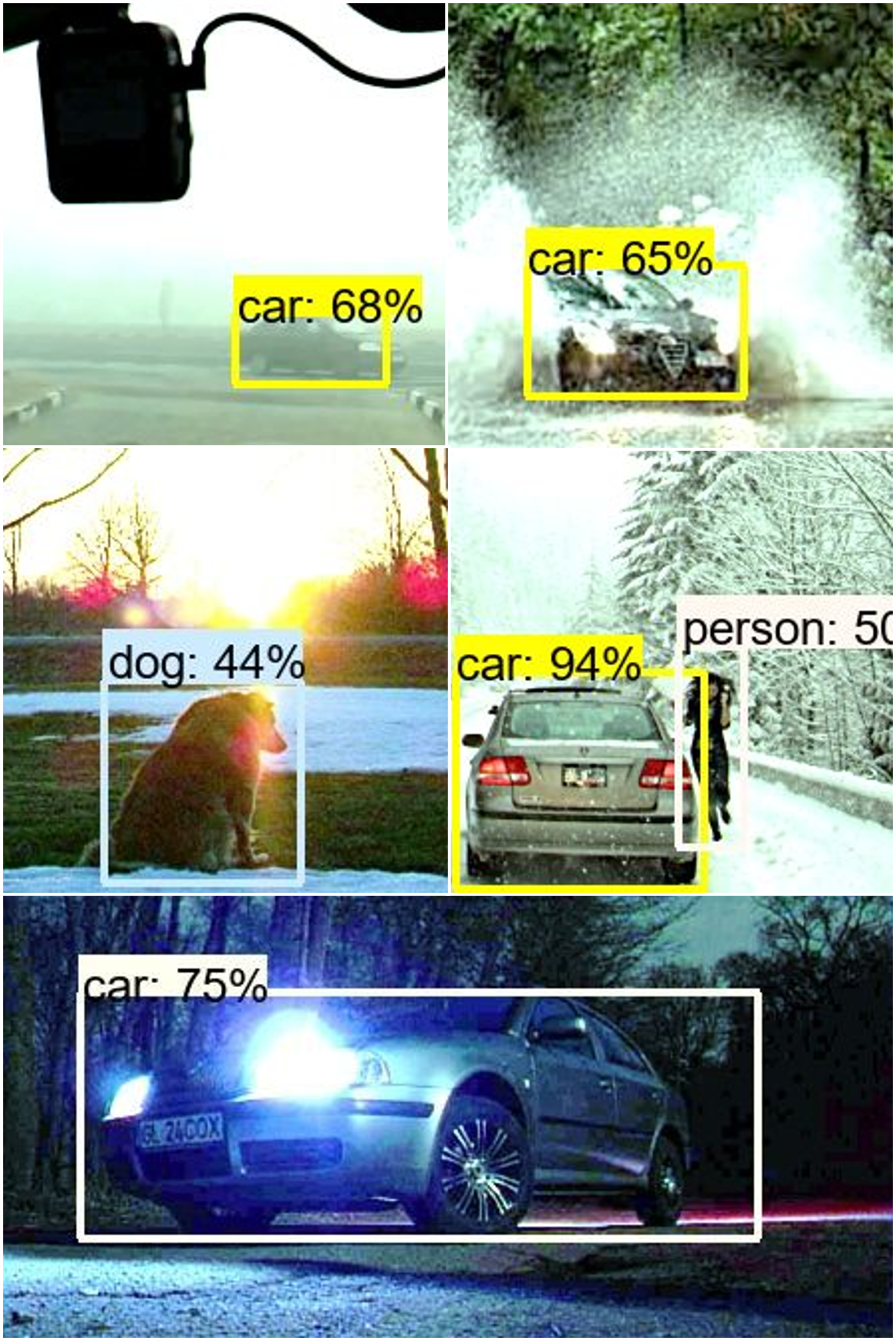}\\
    (a) Adverse weather images&\hspace{-13pt}(b) Proposed method
    \end{tabular}
    \caption{Failure cases of the conventional YOLOv3 detector on adverse weather images (left side) such as fog, rain, low-light and snow conditions, and the successful detection results after applying our proposed image processing filters (right side).}
    \label{fig:imageresults}
\end{figure}

Machine learning-based object detection methods \cite{yolo,yolov3,fasterrcnn,maskrcnn,retinanet,fcos,ssd,carion2020end,zhang2022dino,zhao2024detrs} are typically trained on images captured under normal weather and well-exposed conditions, and have achieved higher performance. However, their performance is degraded significantly when these detectors are applied to adverse weather scenarios such as foggy or low-light conditions. The adverse weather conditions cause domain shifts between the training and testing distributions. To mitigate these domain shifts, several approaches have been proposed. Some methods attempt to reduce the impact of weather by reconstructing the input image with defogging \cite{msbdn,griddehaze}, or low-light enhancement \cite{zerodce}. Another approach\cite{Chen_2018_CVPR,dayolo,priorbased} treats adverse weather as a domain adaptation problem, aiming to align the feature distributions of normal and adverse condition domains. 

Furthermore, an emerging approach called image adaptive object detection that combines image processing and domain adaptation has been proposed. The image-adaptive methods \cite{iayolo, gdipyolo} integrate differentiable image processing filters into the object detection pipeline. These methods estimate the parameters of classical image processing filters such as white balance, gamma correction, tonemapping, contrast enhancement and sharpening from the input image using a Convolutional neural network (CNN) model. The classical filters with the estimated parameters are then applied to the input image as the preprocessing before feeding it to the object detector. 

While these image adaptive method shows promising results, these still have some problems regarding complexity of filter representations. The complexity includes combinations, order, and various parameter range of multiple filters. This complexity leads to redundant filter processing and requires manual customization of filter combinations for each adverse weather condition. As the number of filters increases, the complexity of such customization grows exponentially. Therefore, a simplified representation of preprocessing filters is crucial for achieving an efficient and customization-free image adaptive preprocessor.

This paper proposes a novel image-adaptive object detection method with more simple and customization-free image processing filters.  We reconsidered the role of the classical preprocessing filters, and assumed that these filters boosted object detection performance based global and local enhancement.  Based on this assumption, we unify the classical filters, and represent these filters only by two simple and differentiable filters. Our contributions are as follows:

\begin{itemize}
\item We propose an image-adaptive object detection method, called Enhanced Robustness by Unified Image Processing (ERUP)-YOLO. Our key idea is to unify and generalize classical image processing filters into two simple and differentiable filters: a Bézier curve-based pixel-wise (BPW) filter and a kernel-based local (KBL) filter.
\item Our method does not require data-specific customization of the filter combinations, parameter ranges.
\item Using our BPW filter, we additionally propose a domain-agnostic data augmentation, which can be universally applied to adverse weather conditions.
\end{itemize}
We demonstrate that our proposed BPW and KBL filters match or exceed the expressiveness of the classical filters.  We also show that ERUP-YOLO achieves the highest object detection accuracy based on YOLOv3 compared to state-of-the-art image adaptive methods on adverse weather datasets.

\section{Related Work}
\subsection{Object Detection}
Object detection localizes and classifies objects within images and is a fundamental computer vision task. The main methodologies can be broadly categorized into two groups. The first involves a two-stage process such as FasterRCNN\cite{fasterrcnn} and MaskRCNN\cite{maskrcnn}. The second approach is a one-stage regression-based method such as YOLO series\cite{yolo,yolov3}, RetinaNet\cite{retinanet}, SSD\cite{ssd}, FCOS\cite{fcos}, DETR \cite{carion2020end}, DINO\cite{zhang2022dino} and RT-DETR \cite{zhao2024detrs} which directly predicts the object labels and bounding box coordinates using a single CNN or Transformer based architecture. While these object detection methods have achieved remarkable performance on benchmark datasets, their accuracy often degrades significantly when applied to images captured under adverse conditions, such as fog, haze, or low-light conditions. This is primarily because most conventional detectors are trained on datasets consisting of images acquired under normal weather and well-exposed conditions.

\subsection{Adverse Weather Conditions}
To address the challenge of object detection under adverse weather conditions, various approaches have been explored.  One straightforward approach aims to reconstruct a clear image from an input image degraded by adverse weather conditions \cite{ he2010single, AODnet, griddehaze, bench, msbdn, zerodce}. MSBDN \cite{msbdn} utilizes U-Net architecture for dehazing. GridDehazeNet \cite{griddehaze} introduces an attention mechanism to pre and post-processing to reduce artifacts of a dehazed image. ZeroDCE \cite{zerodce} achieves light enhancement in a no-reference scenario. Alternatively, domain adaptation techniques have been proposed to bridge the domain gap between normal and adverse condition images for the object detection task \cite{Chen_2018_CVPR,dayolo,priorbased, dsnet}. These domain adaptation methods can be categorized into training-based approaches that learn domain-invariant features, and physics-based approaches that leverage physical models of specific conditions. As a domain-invariant approach, DAYOLO \cite{dayolo} performs domain adaptation for YOLO by jointly minimizing object detection and domain classification losses. DSNet\cite{dsnet} improved detection accuracy on foggy datasets by multi-task learning of object detection and image enhancement. Physics-based approach \cite{priorbased} estimated condition-specific priors against rain and haze conditions using physical transmittance models, and employed adversarial training for adaptation. Recently, an emerging approach \cite{iayolo,gdipyolo} for image adaptive object detection has been proposed.

\subsection{Image Adaptive Object Detection}
Image adaptive object detection methods \cite{iayolo,gdipyolo} integrate differentiable image processing filters into the detection pipeline. Liu et al. \cite{iayolo} proposed IA-YOLO, incorporating six filters: defogging, white balance, gamma correction, contrast enhancement, tone mapping, and sharpening. Their method estimates filter parameters from the input image, applies the filters, and feeds the processed image to the detector. While achieving high accuracy on adverse datasets, IA-YOLO requires manually customizing the filter combinations and parameter ranges for different conditions. To address this, Kalwar et al. \cite{gdipyolo} proposed GDIP-YOLO, which processes the six filters in parallel and combines their outputs via a weighted sum. This gating-structured module (GDIP block) eliminates manual customization. They also proposed MGDIP-YOLO, which sequentially processes an input image and its features by using multiple GDIP blocks. These approaches can be interpreted as applying the concept of hypernetworks\cite{chauhan2023brief} to the image processing pipeline. Hypernetworks are neural networks that estimate weight parameters for other neural networks and have been used in image enhancement networks\cite{overparametrization}. Similarly, image adaptive object detection methods estimate parameters for image processing filters, enabling faster and more efficient image transformation for detection tasks.

However, these prior approaches have two remaining limitations. First, independently processing filters may lead to undesired interactions among them and there is potential for improvement. Second, they still require condition-specific data augmentation strategies like applying fog models or gamma adjustments, lacking a unified approach across conditions. To overcome these issues, we aim to unify the functionalities of the filters into an integrated module, while leveraging this unified module for effective and generalized data augmentation. While the image adaptive approach can be applied to various object detection methods, in this study, we follow prior methods and base our evaluation on YOLO.

\begin{figure*}
    \centering
    \begin{tabular}{ccc}
    \includegraphics[height=0.22\vsize]{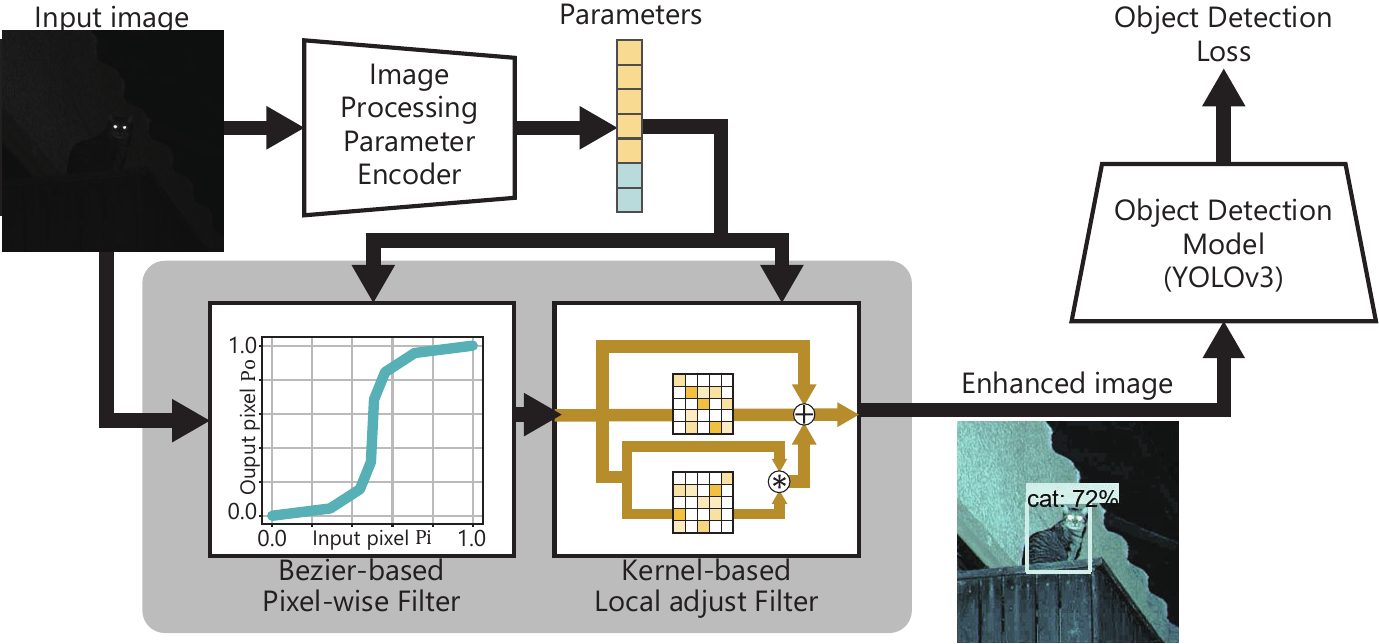}&
    \hspace{0.05\hsize}&
    \includegraphics[height=0.22\vsize]{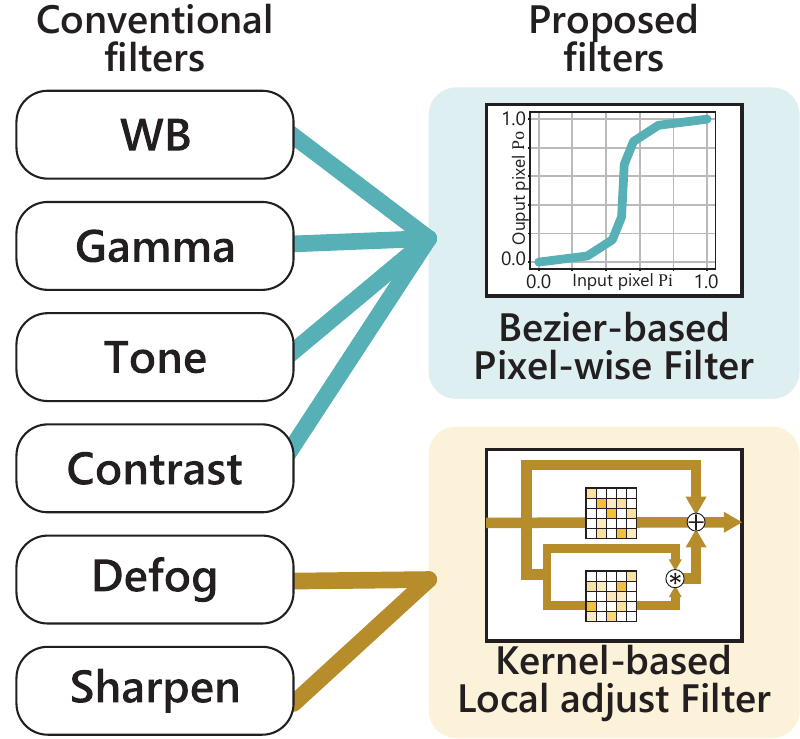}\\
    (a)&&(b)
    \end{tabular}
    \caption{(a) Overview of the proposed ERUP-YOLO framework, which integrates the B\'{e}zier curve-based pixel-wise (BPW) filter and the kernel-based local (KBL) filter before the object detection network. (b) Illustration of how the proposed BPW and KBL filters unify and generalize the conventional image processing filters.}
    \label{fig:framework}
\end{figure*}

\section{Conventional Image Filters}
In this section, we explain the image processing filters used in prior image adaptive methods\cite{iayolo,gdipyolo} to facilitate understanding of our proposed method. Each module requires parameters to determine the degree of processing and is made differentiable. 
Liu et al. \cite{iayolo} proposed these filters and they broadly classify filters into three categories: pixel-wise filters that perform global intensity mapping of pixels, sharpen filters that enhance image sharpness, and defog filters that remove fog.

\subsection{Pixel-wise Filters}
Pixel-wise filters consist of the following four filters: a gamma filter, a white balance filter, a contrast filter, and a tone filter. These filters aim to transform the input image pixel intensity $P_i = (r_i, g_i, b_i)$ to $P_o = (r_o, g_o, b_o)$.

The gamma filter with parameter $\gamma$ and the white balance filter with parameters $(W_r, W_g, W_b)$ are represented by $P_o = P_i^{\gamma}$ and $P_o = (W_r r_i, W_g g_i, W_b b_i)$, respectively. As shown on the left side of Figure\ref{fig:pwfilters}, the gamma filter reduces contrast when the gamma value is less than 1 and enhances low intensities when it is greater than 1.  

The contrast filter transforms intensity values in an S-curve manner as shown on the second figure from the left in Figure\ref{fig:pwfilters}. This filter is expressed using the image luminance $Lum(P_i) = 0.27r_i + 0.67g_i + 0.06b_i$ as follows:
\begin{eqnarray}
    P_o &=& \alpha \cdot En(P_i) + (1-\alpha)\cdot P_i,\\
    En(P_i) &=& P_i \times \frac{0.5(1-\cos(\pi \times Lum(P_i))}{Lum(P_i)}.
\end{eqnarray}
As can be seen from the second figure from the left in Figure\ref{fig:pwfilters}, the parameter $\alpha$ adjusts the shape of the S-curve.

In the tone filter, the range $[0, 1]$ of the input pixel intensity $P_i$ is divided into L intervals, and a piecewise linear function is defined to vary the slope in each interval $[k/L, (k+1)/L]$ for $0 \leq k \leq L-1$.
The filter proposed by Hu et al. \cite{hu2018exposure} has a set of non-negative parameters ${t_0, t_1, ..., t_L}$ corresponding to each interval, and is represented in a differentiable form using $T_L = \Sigma_{l=0}^{L}t_l$ as follows:

\begin{eqnarray}
\label{eq:tone}
P_o = \frac{1}{T_L} \Sigma^{L-1}_{j=0}clip(L \cdot (P_i-j), 0,1)t_j.
\end{eqnarray}
Here, $t_j/T_L$ represents the slope of the line in the interval $[j/L, (j+1)/L]$. In prior implementations \cite{iayolo,gdipyolo}, the values of $t_j$ are set to the range $[0.5, 2]$, and the slope in each interval depends on the relative values of the parameter set. The third figure from left in Figure \ref{fig:pwfilters} shows the extreme parameter cases of the tone filter ($L=8$) where one value of $t_j$ is 0.5 and the others are 2, or vice versa.

\subsection{Sharpen Filter}
The unsharpen mask technique \cite{polesel2000image} is employed to bandpass high-frequency components of the image. This process is expressed using the image $I$, spatial coordinate $x$, Gaussian blur convolution $Gau$, and parameter $\lambda$ as follows:

\begin{eqnarray}
    F(x,\lambda) = I(x) + \lambda(I(x)-Gau(I(x))).
\end{eqnarray}
The prior works \cite{iayolo,gdipyolo} used fixed parameter $\sigma = 5$ and kernel size of 13 for Gaussian blur parameters. This equation can be abstractly represented as the result of convolving the image $I$ with a convolution kernel $K_s$:

\begin{eqnarray}
\label{eq:sharpen}
    F(x,\lambda) &=& Conv(I,K_s),\\
    K_s &=& (1+\lambda)E - \lambda K_{Gau}.
\end{eqnarray}
Here, $E$ is the identity matrix, and $K_{Gau}$ represents the Gaussian kernel.

\subsection{Defog Filter}

Liu et al. \cite{iayolo} proposed a learnable defog filter based on the dehazing method by He et al. \cite{he2010single}. In the basic dehazing method, a hazy image $I(x)$ is formulated as:
\begin{eqnarray}
    I(x) = J(x)t(x)+A(1-t(x)),
\end{eqnarray}
where $t(x)$ is the medium transmission map, and $A$ is the global atmospheric light.
$A$ is estimated from a specific position of each channel of $I$.

Liu et al. \cite{iayolo} introduced a learnable parameter $\omega$ to $t(x)$, defining it as:
\begin{eqnarray}
t(x,\omega) = 1- \omega \min_C{\min_{y \in \Omega(x)}{\frac{I^{C}(y)}{A^{C}}}},
\end{eqnarray}
where $\Omega(x)$ denotes spatial local region around $x$.
Consequently, the dehazed image can be expressed as:
\begin{eqnarray}
J(x) = I(x)\frac{1}{t(x,\omega)} - \frac{A}{t(x,\omega)} + A.
\end{eqnarray}
Here, we attempt to represent this equation in a more abstract form. By letting $F(I,\omega)(x) = 1/t(x,\omega)$, it can be written as:
\begin{eqnarray}
\label{eq:defog}
J(x) = I(x)*F(I,\omega)(x) - A \cdot F(I,\omega)(x) + A,
\end{eqnarray}
where $*$ denotes element-wise multiplication between images.

\subsection{Observation of Conventional Filters} 
Conventional filters require determining the optimal parameter ranges for each filter. For example, in implementations of conventional methods\cite{iayolo,gdipyolo}, the tone filter parameter $t_j$ is specified within the range [0.5, 2], the sharpen filter parameter $\omega$ is set to [0, 5], and the contrast filter parameter $\alpha$ is set to [-1, 1].

We reconsidered the functionalities of the conventional filters and hypothesized that integrating them could lead to a more generalized filter. We view the conventional filters more abstractly: Pixel-wise filters are a set of mappings that transform the input pixel intensity $P_i$ to the output $P_o$, while Sharpen and Defog filters can be represented as combinations of functions that take the image as input. Based on these observations, we aim to unify the filters for simple and effective image enhancement for object detection.

\begin{figure*}
    \centering
        \includegraphics[width=0.95\hsize]{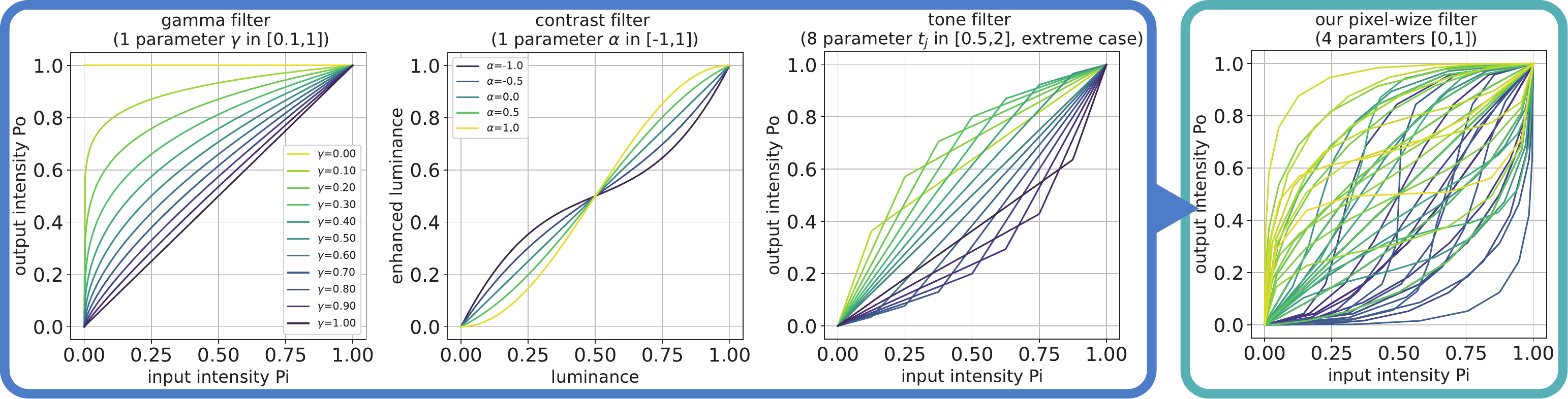}
    \caption{Plots illustrating the input-output pixel intensity mappings of conventional pixel-wise filters (gamma, contrast, tone) compared to the proposed B\'{e}zier curve-based pixel-wise (BPW) filter.}
    \label{fig:pwfilters}
\end{figure*}

\section{Proposed Method}
From the observations on the filters, we propose to employ new filters that unify the functionalities of the conventional filters while retaining the prior framework. Our proposed method follows the prior framework of using a filter-parameter predictor as shown in Figure \ref{fig:framework}(a). The framework applies filters with predicted parameters on an input image, and the filtered image is fed into YOLOv3. The entire framework is trained end-to-end using only detection loss used for YOLOv3 \cite{yolov3}. We introduce two novel differentiable filters as shown in Figure \ref{fig:framework}(b): 1) A B\'{e}zier curve-based pixel-wise (BPW) filter that unifies conventional pixel-wise filters, and 2) A kernel-based local (KBL) filter that comprehensively approximates conventional defog and sharpness filters with local linear functions. Furthermore, leveraging the expressive power of BPW filter, we propose data augmentation using the filter.

\subsection{B\'{e}zier curve-based Pixel-wise Filter}
We chose a cubic Bézier curve for the pixel mapping transform from the input pixel intensity $P_i$ to the output $P_o$. Bézier curves are commonly used for trajectory interpolation, vector data representation, and other applications involving curve modeling. They have also been employed for pixel mapping of low-light image enhancement \cite{bezier1} to improve image visibility. However, utilizing this Bézier curve mapping to make it differentiable with minimal parameters while maintaining low computational cost has not been explored. Therefore, we propose a differentiable Bézier curve-based filter for image-adaptive object detection.

A B\'{e}zier curve is a parametric curve defined by multiple control points and any point on the curve is represented by a linear combination of the control points.

We define the Bezier curve with starting point $[0, 0]$ and ending point $[1, 1]$ expressed as: 

\begin{eqnarray}
C_P(q) = 3q(1-q)^2 p_1 + 3q^2(1-q) p_2 + q^3.
\end{eqnarray}
Here, $q$ is the parametric variable, and $p_1$ and $p_2$ represent the control points $p_j = (P_{ij}, P_{oj})$. Due to the convex hull property, if $p_1$ and $p_2$ are within the range $[0, 1]$, the curve is guaranteed to be contained in $[0, 1]$ and monotonically increasing.

To make the curve differentiable, we define the filter as a B\'{e}zier curve approximating a piecewise linear function used in the tone filter (Eq.\ref{eq:tone}):

\begin{eqnarray}
P_o = \Sigma^{L-1}_{j=0} clip(P_i-C_{P_i}(q_j), 0, \Delta C_{P_i}(q_j))\cdot \frac{\Delta C_{P_o}(q_j)}{\Delta C_{P_i}(q_j)}.
\end{eqnarray}
We divide the parametric space $q \in [0, 1]$ into $L$ intervals and modify the tone filter such that the slope in each interval is $\Delta C_{P_o}(q_j)/\Delta C_{P_i}(q_j)$.

Additionally, we define the filter parameters such that the mapping curve becomes the identity mapping when the parameters are zero:
\begin{eqnarray}
p_1&= [\frac{r_1+1}{2} \cos (\frac{(\theta_1+1)\pi}{4}), \frac{r_1+1}{2} \sin (\frac{(\theta_1+1)\pi}{4})].\\
p_2&= [1-\frac{r_2+1}{2} \cos (\frac{(\theta_2+1)\pi}{4}), 1- \frac{r_2+1}{2} \sin (\frac{(\theta_2+1)\pi}{4})].
\end{eqnarray}
where $\theta_i \in [-1, 1]$ and $r_i \in [-1, 1]$. With this definition, the mapping curve with zero parameters becomes the line passing through $[0, 0]$ and $[1, 1]$. The rightmost of Figure \ref{fig:pwfilters} illustrates the range of curves that our proposed filter can represent, exhibiting higher expressiveness compared to the conventional filters shown in Figure \ref{fig:pwfilters}. We define the mapping process for each RGB channel separately. Therefore, the total number of parameters is 12.

\begin{figure}
    \centering
    \includegraphics[width=0.7\hsize]{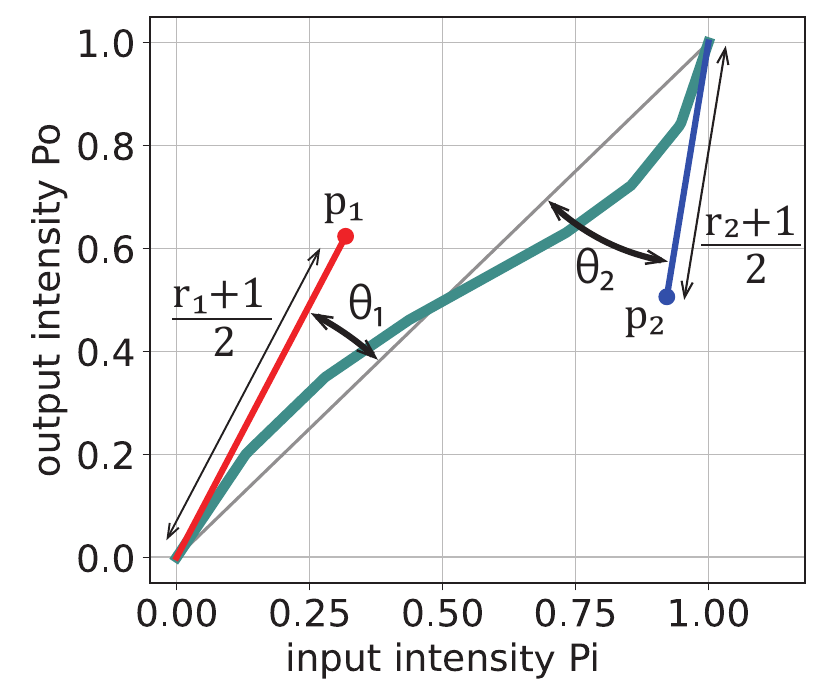}
    \caption{Parameters definition of BPW filter.}
    \label{fig:bezier}
\end{figure}

\subsection{Kernel-based Local Filter}
From implementations of the defog and sharpness filters, we observed that the conventional methods involve combinations of functions that take the image as input. Therefore, we considered integrating each process into a locally linear convolution operation:

\begin{eqnarray}
F_k(I,K_1,K_2) = I * Conv(I,K_1) + Conv(I,K_2) + I.
\end{eqnarray}
Here, $K_j$ represents a convolution kernel, which is a $ksize \times ksize$ matrix. The elements of $K_j$ are constrained to the range $[-1, 1]$. 
$F_k$ approximates and integrates the processing of the conventional defog filter (Eq.\ref{eq:defog}) and sharpen filter (Eq.\ref{eq:sharpen}) into a locally linear convolution operation. The first term of Eq.\ref{eq:defog} corresponds to the first term of this equation, while the remaining terms correspond to the second term. Meanwhile, Eq.\ref{eq:sharpen} corresponds to the second term of this equation.

Similar to BPW filter, we apply this filter to each image channel with separate parameters. Therefore, the total number of parameters is $2 \times ksize^2 \times 3$, and we set $ksize = 9$.

\begin{figure*}
    \centering
        \includegraphics[width=0.85\hsize]{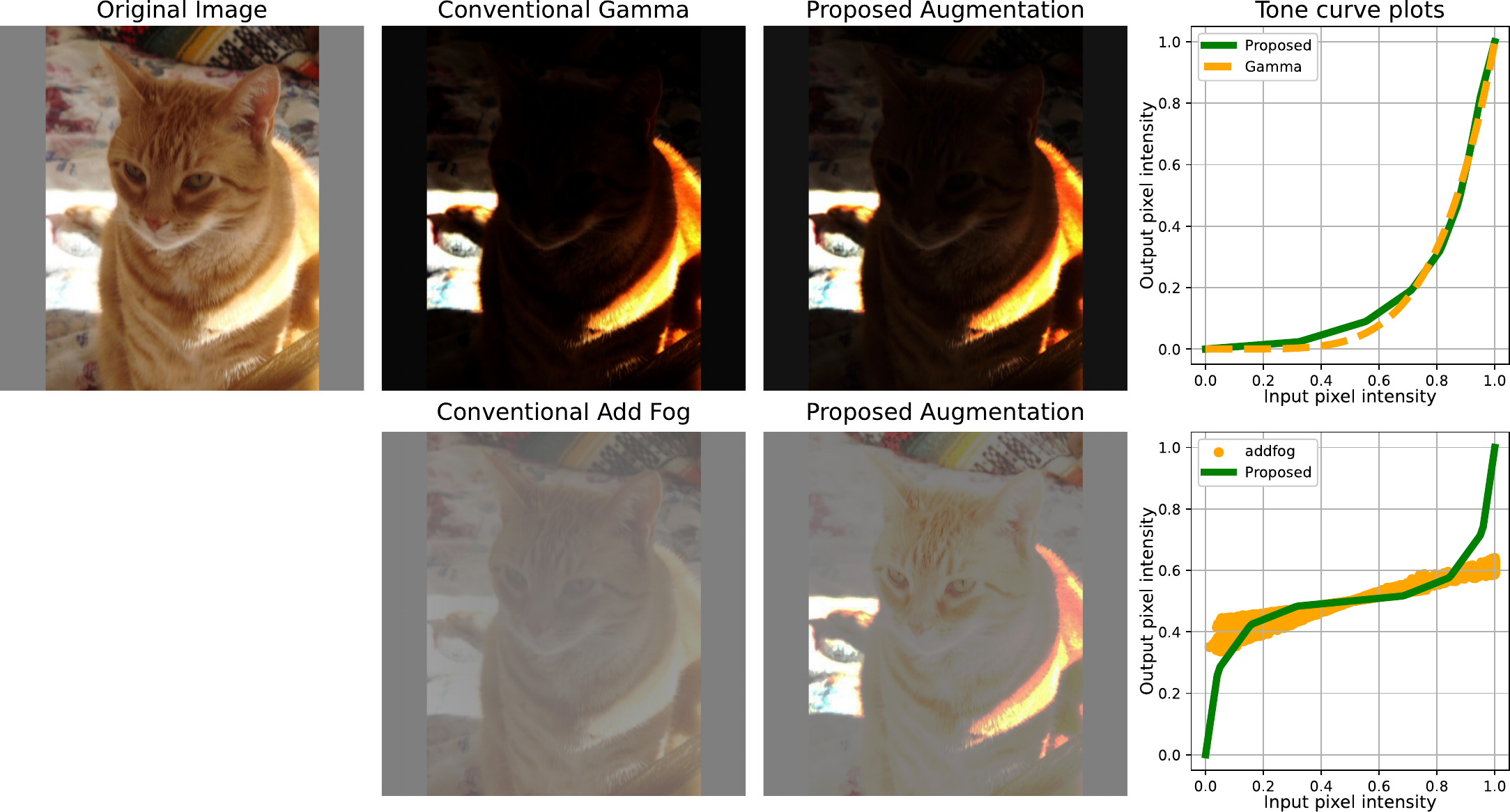}
    \caption{Comparison of prior data-specific augmentation (2nd column) and the proposed B\'{e}zier curve-based pixel-wise (BPW) augmentation (3rd column) for low-light (top row) and foggy (bottom row) conditions. The 3rd column shows BPW augmentation manually adjusted to mimic the prior augmentations. The rightmost column visualizes the pixel intensity mappings as plots.}
    \label{fig:augment}
\end{figure*}

\subsection{Image Processing Parameter Encoder}
Following the framework of the high-performing GDIP-YOLO \cite{gdipyolo} from conventional research, we define an encoder with the following structure. The encoder comprises five convolutional layers, each with a kernel size of 3 and a stride of 1. The number of channels starts from 64 and doubles after each layer, with the final layer having 1024 channels. Each convolutional layer is followed by average pooling with a kernel size of 3 and a stride of 2. After the final layer, global average pooling is applied, resulting in an output of $1 \times 1 \times 1024$. The output vector is then projected to a latent space with dimensions equal to the parameter size using a fully connected layer. The projected vector is passed through a sigmoid function to constrain the values between 0 and 1. The parameters in this vector are then scaled and mapped to the appropriate range for each filter.  

\subsection{BPW Filter for Data Augmentation}
\label{seq:bpwaug}
In addition to image enhancement, our BPW filter enables its use for data augmentation during training by applying diverse intensity mappings to the input images. Prior methods employed domain-specific augmentations, using physical fog models for foggy datasets and gamma adjustments for low-light datasets. In contrast, our approach can adapt to both fog and low-light conditions using only the proposed BPW filter. Figure \ref{fig:augment} illustrates how our filter can manually mimic the appearance of images augmented by the prior dataset-specific methods through appropriate parameter settings.

\section{Experiments}
Following prior image adaptive methods, we first evaluate our method using separate models trained for foggy and low-light conditions with a different number of object classes. We then assess the performance of a single model across a diverse range of adverse weather conditions, including fog, low-light, rain, snow, and sand. Finally, we conduct an ablation study to validate the effectiveness of each proposed module in our framework.

\subsection{Experimental Settings of Object Detection}
\label{seq:expset}
We evaluate the object detection accuracy of our method and prior method from different categories: image reconstruction \cite{msbdn,griddehaze,zerodce}, domain adaptation \cite{dayolo}, multi-task learning \cite{dsnet}, and image adaptive object detection \cite{iayolo,gdipyolo}. For fair comparisons, we employ the YOLOv3 \cite{yolov3} as the common backbone for object detection across all methods. All models are trained on the clear images from VOC 2007/2012 trainval sets \cite{vocdataset}, with data augmentation applied to simulate adverse weather conditions. Note that no real-world adverse weather data is used during training.

We evaluate the object detection performance using a mean Average Precision (mAP) metric at an Intersection over Union (IoU) threshold of 0.5. All detection models are trained for 80 epochs using the SGD optimizer with a cosine learning rate scheduler (initial rate of $10^{-4}$ decaying to $10^{-6}$) and a weight decay of $5 \times 10^{-4}$. The input image resolution is set to $448 \times 448$, with a batch size of 6.

\subsection{Evaluation on Foggy Conditions} 
For training on foggy conditions, our proposed approach leverages BPW filter for data augmentation, applying the filter with random parameters to the input images during training. Note that we follow the setting of data-augmentation \cite{narasimhan2002vision} used for prior methods\cite{iayolo,gdipyolo}.
We evaluate our approach on three datasets: the PASCAL VOC 2007 test set \cite{vocdataset} (V\_n\_ts), the synthesized VOC\_Foggy\_test dataset (V\_f\_ts) \cite{iayolo} generated by applying the atmospheric scattering model \cite{narasimhan2002vision} to the PASCAL VOC 2007 test set, and the RTTS dataset \cite{bench} containing real-world foggy images. As conventional image reconstruction methods for foggy images, MSBDN \cite{msbdn} and GridDehaze \cite{griddehaze} are used for evaluation.

Table \ref{tb:res_fog} shows the results on the three datasets. Our proposed method achieves the highest performance on all datasets. These results demonstrate the effectiveness of our unified filtering strategy in enhancing object detection under foggy conditions.

\begin{table}[t]
    \centering
    \tiny
    \caption{Results of mAP50 on foggy conditions using 5 class models. Best and second scores are bold and underlined.}
    \label{tb:res_fog}
    \begin{tabular}{llcccc}
    \\
    \multirow{2}{*}{{\scriptsize Methods}}&&{\tiny Train data}&{\scriptsize V\_n\_ts}&{\scriptsize V\_f\_ts}&{\scriptsize RTTS}\\
    &&{\tiny Augment}&&&\\
    \\
    \hline
    \hline
    \multirow{3}{*}{{\scriptsize Baseline}}
    \\
    &{\scriptsize YOLOv3\cite{yolov3}}&{\footnotesize fog}&{\footnotesize 64.13}&{\footnotesize 63.40}&{\footnotesize 30.80}\\
    \\
    \hline
    \\
    \multirow{3}{*}{\scriptsize Defog}&{\scriptsize MSBDN\cite{msbdn}}&{\footnotesize -}&{\footnotesize -}&{\footnotesize 57.38}&{\footnotesize 30.20}\\
    \\
    &{\scriptsize GridDehaze\cite{griddehaze}}&{\footnotesize -}&{\footnotesize -}&{\footnotesize 58.23}&{\footnotesize 31.42}\\
    \\
    \hline
    \multirow{3}{*}{
    \hspace{-3mm}
    \begin{tabular}{l}
    {\scriptsize Domain}\\
    {\scriptsize Adaptation}
    \end{tabular}
    }
    \\
    &{\scriptsize DAYOLO\cite{dayolo}}&{\footnotesize fog}&{\footnotesize 56.51}&{\footnotesize 55.11}&{\footnotesize 29.93}\\
    \\
    \hline
    \\
    {\scriptsize Multi-task}&{\scriptsize DSNet\cite{dsnet}}&{\footnotesize fog}&{\footnotesize 53.29}&{\footnotesize 67.40}&{\footnotesize 28.91}\\
    \\
    \hline
    \\
    \multirow{5}{*}{
    \hspace{-3mm}
    \begin{tabular}{l}
    {\scriptsize Image}\\
    {\scriptsize Adaptive}
    \end{tabular}
    }
    &{\scriptsize IA-YOLO\cite{iayolo}}&{\footnotesize fog}&{\footnotesize 73.23}&{\footnotesize 72.03}&{\footnotesize 37.08}\\
    &{
    \multirow{3}{*}{
    \hspace{-3mm}
    \begin{tabular}{l}
    {\scriptsize GDIP}\\
    {\scriptsize -YOLO\cite{gdipyolo}}
    \end{tabular}
    }
    }\\
    &&{\footnotesize fog}&{\footnotesize 73.70}&{\footnotesize 71.92}&{\footnotesize 42.42}\\
    &{
    \multirow{3}{*}{
    \hspace{-3mm}
    \begin{tabular}{l}
    {\scriptsize MGDIP}\\
    {\scriptsize -YOLO\cite{gdipyolo}}
    \end{tabular}
    }
    }\\
    &
    &{\footnotesize fog}&{\footnotesize \underline{75.36}}&{\footnotesize \underline{73.37}}&{\footnotesize \underline{42.84}}\\
    \\
    \hline
    \\
    {\scriptsize Proposed}&{\scriptsize ERUP-YOLO}&{\footnotesize BPW}&{\footnotesize {\bf 77.89}}&{\footnotesize {\bf 74.09}}&{\footnotesize {\bf 49.81}}\\
    \\
    \end{tabular}
\end{table}

\subsection{Evaluation on Low-Light Conditions} 
For training on low-light scenarios, similar to the foggy case, our proposed approach leverages BPW filter for data augmentation. Note that we follow the setting of gamma-based data-augmentation used for prior methods\cite{iayolo,gdipyolo}.
We evaluate our method on three low-light datasets: the PASCAL VOC 2007 test set \cite{vocdataset} (V\_n\_ts), the synthesized VOC\_Dark\_test set (V\_d\_ts) \cite{iayolo} generated by applying random gamma correction to the PASCAL VOC 2007 test set, and the real-world ExDark dataset \cite{exdark}. As conventional image reconstruction methods for foggy images, Zero-DCE\cite{zerodce} is used for evaluation.

The mAP results on the low-light datasets are presented in Table \ref{tb:res_ll}. Our proposed approach achieves the highest performance on the ExDark dataset and the V\_d\_ts dataset. On the synthetic V\_n\_ts dataset, our method attains the second-best performance.

\begin{table}[t]
    \centering
    \tiny
    \caption{Results of mAP50 on low-light conditions using 10 class models. Best and second scores are bold and underlined.}
    \label{tb:res_ll}
    \begin{tabular}{llcccc}
    \\
    \multirow{2}{*}{{\scriptsize Methods}}&&{\tiny Train data}&{\scriptsize V\_n\_ts}&{\scriptsize V\_d\_ts}&{\scriptsize ExDark}\\
    &&{\tiny Augment}&&&\\
    \\
    \hline
    \hline
    \multirow{3}{*}{{\scriptsize Baseline}}
    \\
    &{\scriptsize YOLOv3\cite{yolov3}}&{\scriptsize gamma}&{\footnotesize 65.33}&{\footnotesize 52.28}&{\footnotesize 37.03}\\
    \\
    \hline
    \multirow{3}{*}{{\scriptsize Enhance}}
    \\
    &{\scriptsize ZeroDCE\cite{zerodce}}&{\footnotesize -}&{\footnotesize -}&{\footnotesize 33.57}&{\footnotesize 34.41}\\
    \\
    \hline
    \multirow{3}{*}{
    \hspace{-3mm}
    \begin{tabular}{l}
    {\scriptsize Domain}\\
    {\scriptsize Adaptation}
    \end{tabular}
    }
    \\
    &{\scriptsize DAYOLO\cite{dayolo}}&{\scriptsize gamma}&{\footnotesize 41.68}&{\footnotesize 21.53}&{\footnotesize 18.15}\\
    \\
    \hline
    \multirow{3}{*}{{\scriptsize Multi-task}}
    \\
    &{\scriptsize DSNet\cite{dsnet}}&{\scriptsize gamma}&{\footnotesize 64.08}&{\footnotesize 43.75}&{\footnotesize 36.97}\\
    \\
    \hline
    \\
    \multirow{5}{*}{
    \hspace{-3mm}
    \begin{tabular}{l}
    {\scriptsize Image}\\
    {\scriptsize Adaptive}
    \end{tabular}
    }
    &{\scriptsize IA-YOLO\cite{iayolo}}&{\scriptsize gamma}&{\footnotesize {\bf 70.02}}&{\footnotesize \underline{59.40}}&{\footnotesize 40.37}\\
    &{
    \multirow{3}{*}{
    \hspace{-3mm}
    \begin{tabular}{l}
    {\scriptsize GDIP}\\
    {\scriptsize -YOLO\cite{gdipyolo}}
    \end{tabular}
    }
    }\\
    &&{\scriptsize gamma}&{\footnotesize 63.23}&{\footnotesize 57.85}&{\footnotesize \underline{42.56}}\\ 
    &{
    \multirow{3}{*}{
    \hspace{-3mm}
    \begin{tabular}{l}
    {\scriptsize MGDIP}\\
    {\scriptsize -YOLO\cite{gdipyolo}}
    \end{tabular}
    }
    }
    \\
    &&{\scriptsize gamma}&{\footnotesize 62.86}&{\footnotesize 57.91}&{\footnotesize 40.96}\\
    \\
    \hline
    \\
    {\scriptsize Proposed}&{\scriptsize ERUP-YOLO}&{\footnotesize BPW}&{\footnotesize \underline{68.62}}&{\footnotesize \bf{59.81}}&{\footnotesize {\bf 48.43}}\\
    \\
    \end{tabular}
\end{table}

\begin{figure}[t]
   \centering
   \begin{tabular}{ccc}
   \includegraphics[width=0.33\hsize]{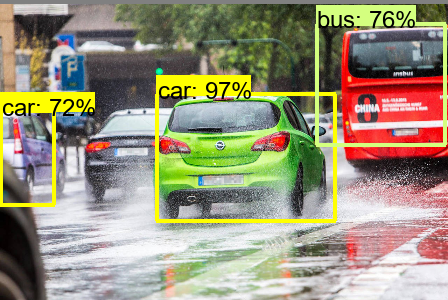}&
   \hspace{-10pt}\includegraphics[width=0.33\hsize]{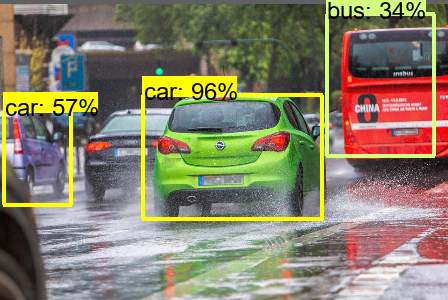}&
   \hspace{-10pt}\includegraphics[width=0.33\hsize]{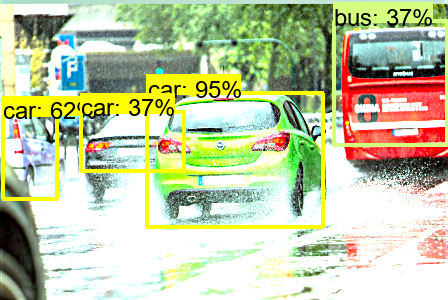}
   \\
   \includegraphics[width=0.33\hsize]{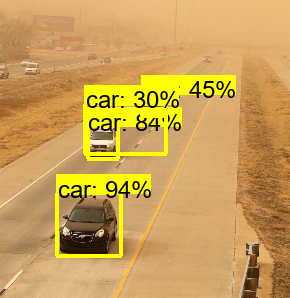}&
   \hspace{-10pt}\includegraphics[width=0.33\hsize]{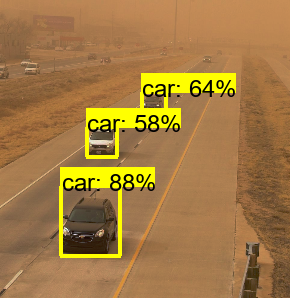}&
   \hspace{-10pt}\includegraphics[width=0.33\hsize]{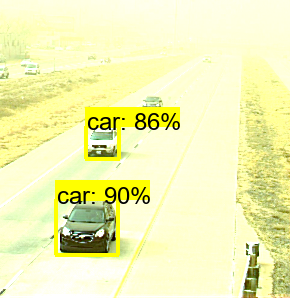}
   \\
   (a)YOLOv3 \cite{yolov3} &\hspace{-10pt}(b)GDIP \cite{gdipyolo} &\hspace{-10pt}(c)ERUP(ours)\\
   \end{tabular}
   \caption{Object detection results (confidence $\geq$ 0.3) on rain and sand images from DAWN dataset \cite{dawn}}
   \label{fig:fail}
\end{figure}

\subsection{Evaluation on Adverse Weather Conditions} 
\label{seq:advweather}
To evaluate the robustness of our proposed method across various adverse weather conditions, we conducted experiments comparing our approach with GDIP-YOLO\cite{gdipyolo} and YOLOv3\cite{yolov3}. All models were trained using the data augmentation strategy with our BPW module, as described in Section \ref{seq:bpwaug} and training method is same in Section \ref{seq:expset}.
We evaluated the models on four datasets: VOC test, ExDark, RTTS, and the DAWN dataset \cite{dawn}. The DAWN dataset comprises 1000 images capturing diverse adverse weather conditions including fog, sand, rain, and snow. For this experiment, we evaluate five common object classes across all datasets and the object detection models are trained using these classes: 'person', 'bicycle', 'car', 'bus', and 'motorcycle (motorbike)'.

\begin{table*}[h]
    \centering
    \small
    \caption{Results of mAP50 on 6 adverse weather conditions using 5 class models trained with same manner.}
    \label{tb:res_real}
    \begin{tabular}{l|ccccccc}
    &VOC\cite{vocdataset}&ExDark\cite{exdark}&RTTS\cite{bench}&\multicolumn{4}{c}{DAWN\cite{dawn}}\\
    &normal&lowlight&fog&sand&rain&fog&snow\\
    \hline 
    YOLOv3\cite{yolov3}&76.38&47.95&49.01&{\bf 31.18}&31.35&32.26&48.82\\
    GDIP-YOLO\cite{gdipyolo}&76.65&47.07&47.81&29.82&29.96&31.66&59.03\\
    ERUP-YOLO(ours)&{\bf 77.89}&{\bf 49.54}&{\bf 49.81}&27.82&{\bf 32.00}&{\bf 32.55}&{\bf 59.82}\\
    \hline
    \end{tabular}
\end{table*}

\begin{figure*}[t]
   \centering
   \includegraphics[width=0.85\hsize]{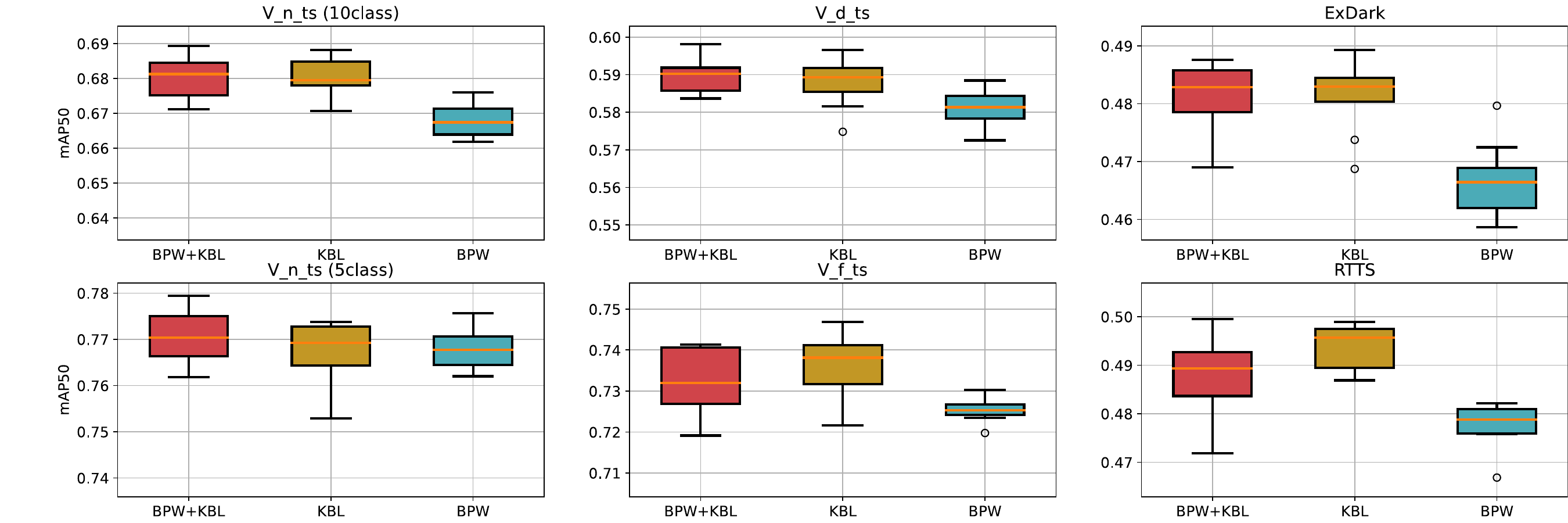}
   \caption{Ablation study on the effects of the proposed BPW and KBL filters across different datasets. Among the datasets, the model combining both BPW and KBL filters achieves the highest performance in most cases.}
   \label{fig:ablationplots}
\end{figure*}

Table \ref{tb:res_real} presents the mAP50 for each model across six weather conditions on the four datasets. Our proposed method consistently outperformed other methods in most weather conditions except sand. Figure \ref{fig:fail} shows detection results examples for rain and sand conditions. Our method demonstrates well performance in rainy scenarios. However, in case of the sand condition, our method lead to over-amplification of pixel values in distant areas that are originally brighter due to atmospheric scattering. 

\begin{figure}[t]
   \centering
   \begin{tabular}{cc}
   \includegraphics[width=0.45\hsize]{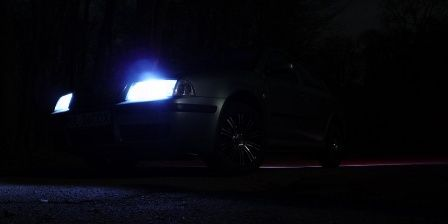}&
   \hspace{-10pt}\includegraphics[width=0.45\hsize]{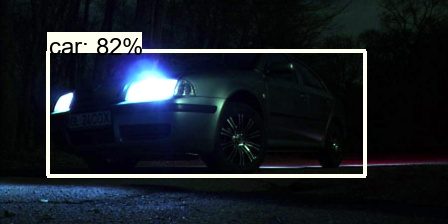}\\
   (a) Original&\hspace{-10pt}(b) KBL\\
   \includegraphics[width=0.45\hsize]{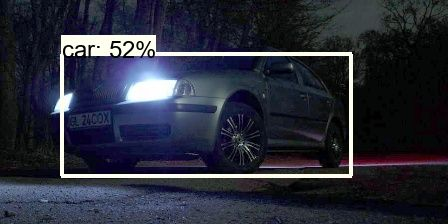}&
   \hspace{-10pt}\includegraphics[width=0.45\hsize]{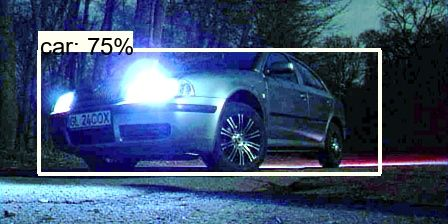}\\
   (c) BPW&\hspace{-10pt}(d) BPW+KBL
   \end{tabular}
   \caption{Visual examples showing the effects of the proposed filters on an input image. The proposed filters (d) show the best visual quality in three settings. }
   \label{fig:ablationresults}
\end{figure}

\subsection{Ablation Study}
We evaluate the individual contributions of the proposed BPW and KBL filters using detection performance on six datasets. We trained each model configuration 8 times as described in Section \ref{seq:expset} and analyzed the distribution of mAP50 scores using box plots, as shown in Figure \ref{fig:ablationplots}. Among the six plots, the model combining both BPW and KBL filters achieves the highest performance in most cases. The model with only KBL filter also demonstrates competitive accuracy.

We noticed that in some foggy conditions, BPW+KBL combination yields slightly lower performance compared to KBL alone. This performance drop can be attributed to the function of BPW filter to globally increase image brightness. In foggy scenes, pixels in distant areas already have high pixel values due to scattering. The function of BPW filter to increase global brightness can lead to oversaturation in these areas. This phenomenon is similar to the results observed in sand conditions, as discussed in Section \ref{seq:advweather}. On the other hand, in normal and low-light conditions (V\_n\_ts, V\_d\_ts and ExDark), BPW+KBL combination is better than KBL or BPW. As illustrated in Figure \ref{fig:ablationresults}, BPW+KBL effectively enhances the brightness of objects in dark areas than others. KBL shows better results in object detection, but BPW+KBL combination provides superior visibility enhancements. We provide additional examples on the supplementary material.

\section{Conclusion}
In this paper, we proposed an image-adaptive object detection method, called ERUP-YOLO with unified and customization-free image processing filters: BPW and KBL filters. Our method does not require data-specific customization of the filter combinations and parameter ranges. We additionally proposed a domain-agnostic data augmentation strategy using our BPW filter, which can be universally applied to multiple weather conditions. We confirmed that our unified filtering approach achieves state-of-the-art object detection performance under foggy and low-light conditions and demonstrate our method can be generalized other conditions.

{\small
\bibliographystyle{ieee_fullname}
\bibliography{reference}
}

\maketitlesupplementary
\section{Visual Comparison with Prior Methods}
We present a visual comparison of detection results between our proposed method and prior approaches (YOLOv3 and GDIP-YOLO) across various adverse weather conditions in Section 5.4 of the main paper. We use the models evaluated in Table 3. Figure \ref{fig:conv} shows enhanced input image and detection results overlaid on input images for low-light, foggy, and rain conditions from ExDark and DAWN datasets. Figure \ref{fig:conv2} extends this comparison to snow and sand conditions.

Our proposed method demonstrates a significant image transformation effect compared to GDIP-YOLO and notable improvements in detection performance, particularly in low-light and snow conditions (first row in Figure \ref{fig:conv} and first row in Figure \ref{fig:conv2}). However, we observe limitations in foggy and sand conditions, particularly for distant objects obscured by scattering effects. In these cases (Figure \ref{fig:conv}, second row for foggy, and Figure \ref{fig:conv2}, second and third row for sand), our method tends to overemphasize the brightness of distant objects affected by scattering, sometimes leading to missed detections in the background of the image.

\section{Visual Results of Ablation Study}
We present a visual comparison of detection results from our Ablation Study in Section 5.5 of the main paper. Figure \ref{fig:abl} presents detection results for low-light and foggy conditions using different filter configurations. BPW+KBL combination consistently improves the overall visibility of the scenes across both low-light and foggy conditions. 
As evident in the second row of Figure \ref{fig:abl}, BPW+KBL configuration significantly enhances the detection accuracy of objects in extremely dark regions. However, the second and third rows also reveal a limitation of BPW+KBL combination. The global brightness increase tends to oversaturate areas that are already bright due to illumination sources or distant scattering effects in foggy conditions. 

\section{Evaluation of Image Filters}
To quantitatively assess the capabilities of our proposed filters, we conduct four sets of experiments on VOC2007\_test dataset. We evaluate the expressiveness of our BPW filter in approximating sequential prior pixel-wise filters (white balance, gamma, contrast, and tone filters). We apply these pixel-wise filters sequentially with random parameters to the test images to generate degraded target images. Then, we optimize the parameters of our BPW filter using the Adam optimizer for 50 iterations with a learning rate of 0.01, minimizing the combined mean squared error (MSE) and structural similarity (SSIM) loss between the filtered images and the target degraded images.

We also conduct the reverse experiment, where our BPW filter randomly degrades the test images, and we measure how well the PW filters can recover the original images. For our KBL filter, we follow a similar procedure. We degrade the test images by applying the combination of prior defog and sharpen filters with random parameters, creating the target degraded images. We also perform the reverse experiment between the defog and sharpen filters and KBL filter.

Table \ref{tb:mimic} shows the PSNR between the degraded and optimized images when using each filter for degradation and optimization, respectively. Since the degree of degradation from the original image varies depending on the degradation filter, we report the mean of the initial PSNR between the original and degraded images, the final PSNR between the optimized and degraded images, and the increase in PSNR (det) from initial to final. The results demonstrate that our BPW filter can mimic the effects of prior pixel-wise filters with comparable performance, while our KBL filter exhibits superior expressiveness in approximating the combination of prior sharpness and defog filters.

The pixel-wise filters shows a slightly higher mimicking capability compared to our BPW filter. This can be attributed to the fact that the pixel-wise filters comprises a sequential process of four distinct pixel-wise filters, allowing for a high degree of expressiveness through their combination. In contrast, our method achieves nearly equivalent expressiveness using only a single module, which is an improved the conventional tone filter. Despite this minor difference, our approach offers significant advantages. As illustrated in Figure \ref{fig:conv}, \ref{fig:conv2}, \ref{fig:abl}, our proposed filter demonstrates high expressiveness in image transformation.

\begin{table}[t]
    \centering
    \small
    \caption{PSNR between the degraded and optimized images when using each filter. Each column presents the initial degraded image (initial), the result after applying the optimization filter (final), and the improvement from initial to final (del).}
    \label{tb:mimic}
    \begin{tabular}{ll|ccc}
    \multicolumn{2}{c|}{Image filters}&\multicolumn{3}{c}{PSNR}\\ 
    Degrade&Optimize&Initial&Final&Det\\
    \hline
    Pixel Wise&BPW(ours)&13.90&22.29&8.38\\
    BPW(ours)&Pixel Wise&17.62&27.46&{\bf 9.84}\\    
    \hline
    Sharp+DeFog&KBL(ours)&12.57&20.09&{\bf 7.52}\\
    KBL(ours)&Sharp+DeFog&-8.17&-8.00&0.17\\
    \end{tabular}
\end{table}

\newpage
\def\imsize{0.3\hsize}

\begin{figure*}[t]
   \centering
   \begin{tabular}{ccc}
   \includegraphics[width=\imsize]{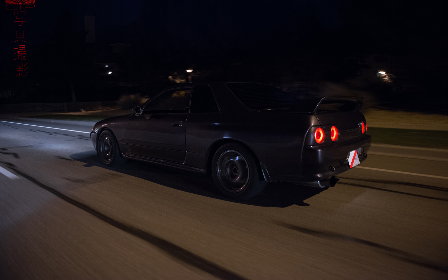}&
   \hspace{-10pt}\includegraphics[width=\imsize]{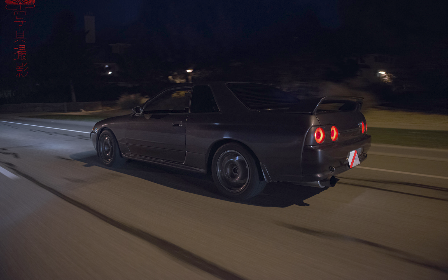}&
   \hspace{-10pt}\includegraphics[width=\imsize]{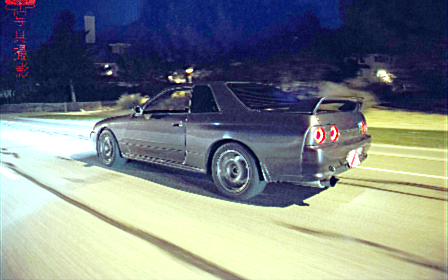}\\ \vspace{-15pt} \\
   \includegraphics[width=\imsize]{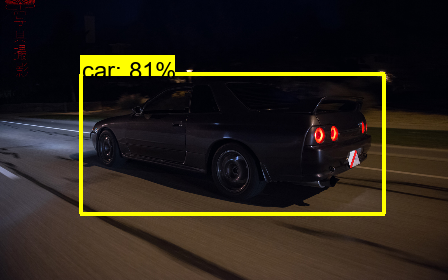}&
   \hspace{-10pt}\includegraphics[width=\imsize]{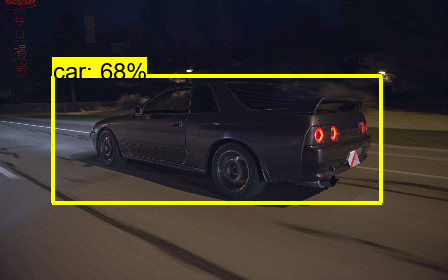}&
   \hspace{-10pt}\includegraphics[width=\imsize]{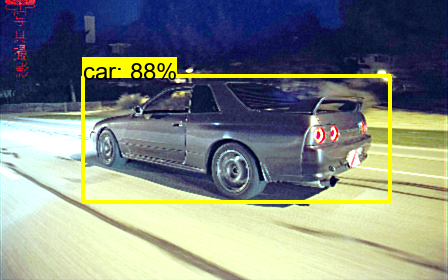}   
    \\
   \includegraphics[width=\imsize]{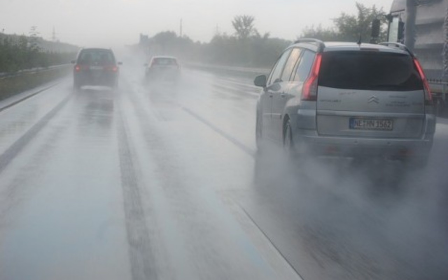}&
   \hspace{-10pt}\includegraphics[width=\imsize]{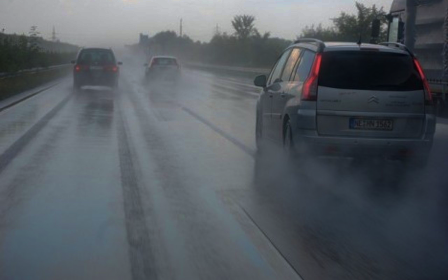}&
   \hspace{-10pt}\includegraphics[width=\imsize]{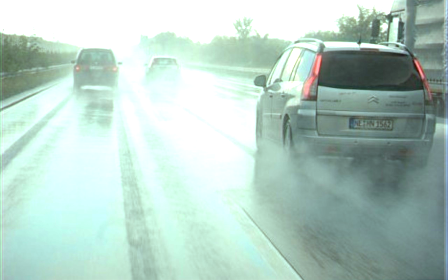}\\ \vspace{-15pt} \\
   \includegraphics[width=\imsize]{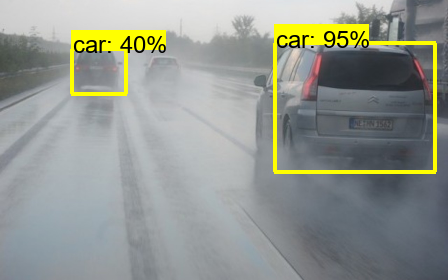}&
   \hspace{-10pt}\includegraphics[width=\imsize]{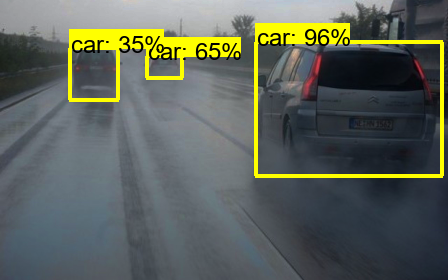}&
   \hspace{-10pt}\includegraphics[width=\imsize]{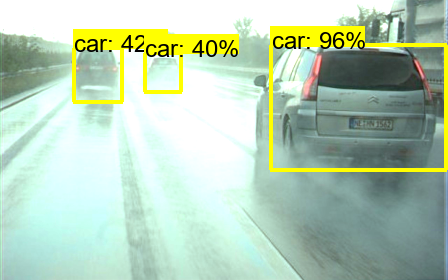}\\
   \includegraphics[width=\imsize]{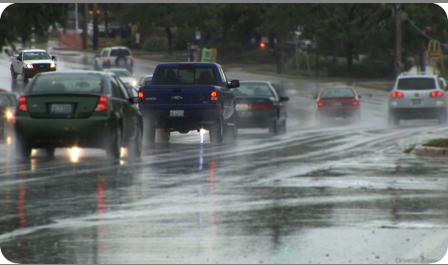}&
   \hspace{-10pt}\includegraphics[width=\imsize]{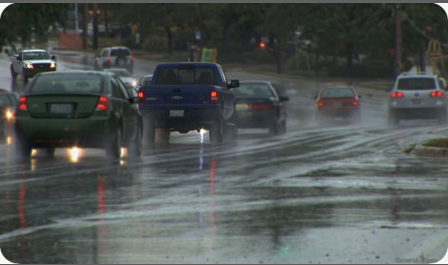}&
   \hspace{-10pt}\includegraphics[width=\imsize]{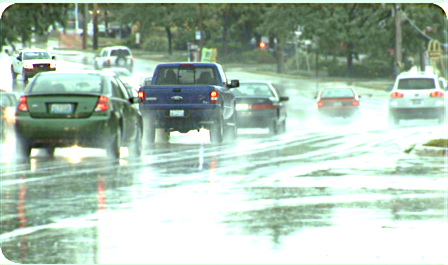}\\ \vspace{-15pt} \\
   \includegraphics[width=\imsize]{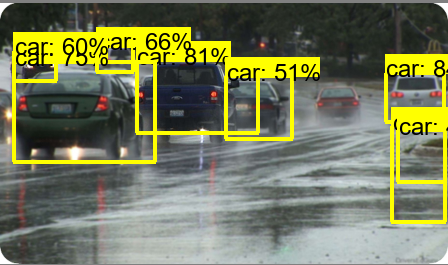}&
   \hspace{-10pt}\includegraphics[width=\imsize]{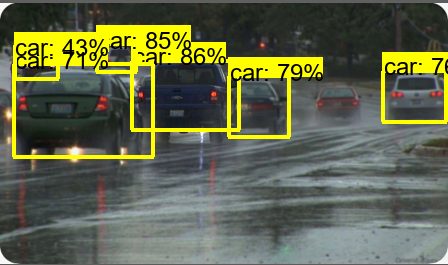}&
   \hspace{-10pt}\includegraphics[width=\imsize]{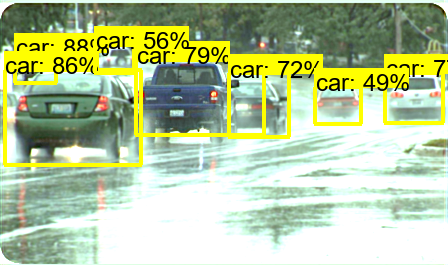}\\
   (a)YOLOv3 &\hspace{-10pt}(b)GDIP &\hspace{-10pt}(c)ERUP(ours)\\
   \end{tabular}
   \caption{Visual comparison of enhanced images and detection results using (a)YOLOv3, (b)GDIP and (c)ERUP(ours). First row shows the low-light results from ExDark, second row shows the fog results from DAWN, and third row shows the rain results from DAWN.}
   \label{fig:conv}
\end{figure*}

\newpage
\def\imsize{0.3\hsize}

\begin{figure*}[t]
   \centering
   \begin{tabular}{ccc}
   \includegraphics[width=\imsize]{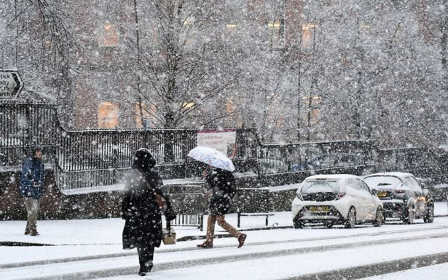}&
   \hspace{-10pt}\includegraphics[width=\imsize]{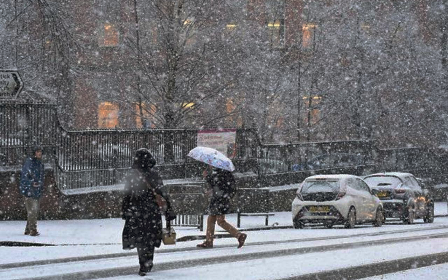}&
   \hspace{-10pt}\includegraphics[width=\imsize]{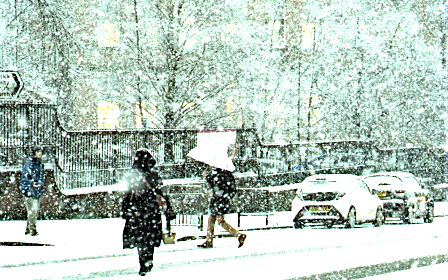}\\ \vspace{-15pt} \\
   \includegraphics[width=\imsize]{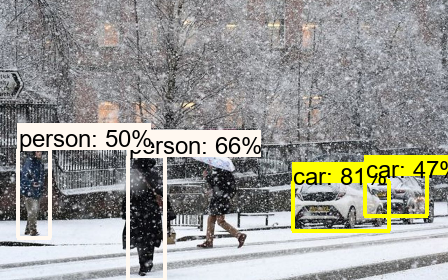}&
   \hspace{-10pt}\includegraphics[width=\imsize]{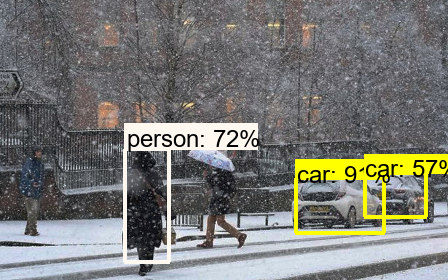}&
   \hspace{-10pt}\includegraphics[width=\imsize]{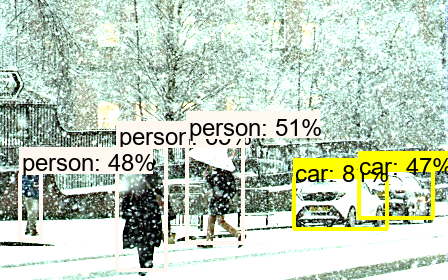}\\
   \includegraphics[width=\imsize]{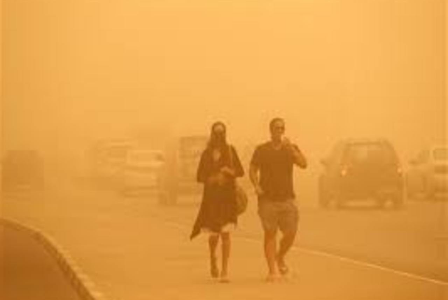}&
   \hspace{-10pt}\includegraphics[width=\imsize]{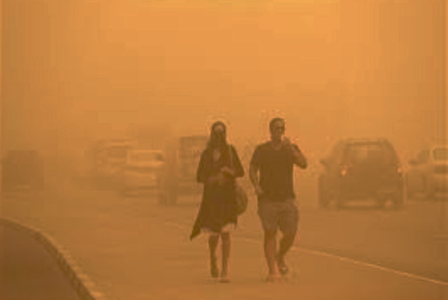}&
   \hspace{-10pt}\includegraphics[width=\imsize]{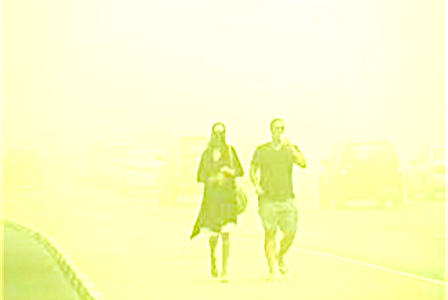}\\ \vspace{-15pt} \\
   \includegraphics[width=\imsize]{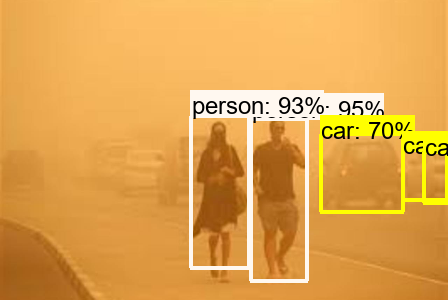}&
   \hspace{-10pt}\includegraphics[width=\imsize]{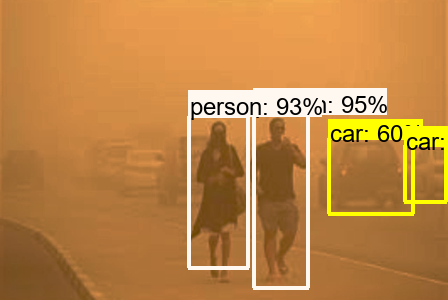}&
   \hspace{-10pt}\includegraphics[width=\imsize]{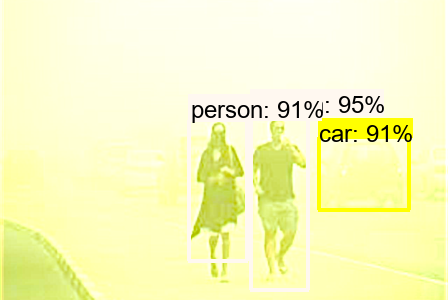}\\
   \includegraphics[width=\imsize]{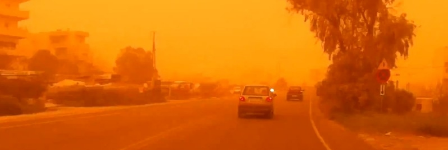}&
   \hspace{-10pt}\includegraphics[width=\imsize]{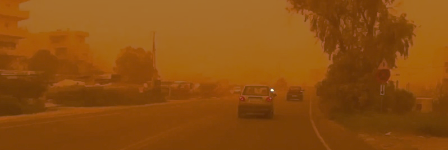}&
   \hspace{-10pt}\includegraphics[width=\imsize]{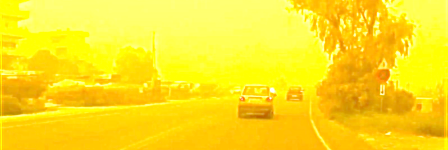}\\ \vspace{-15pt} \\
   \includegraphics[width=\imsize]{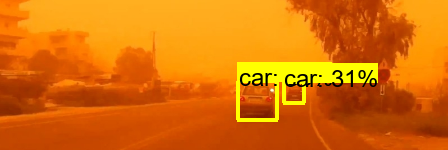}&
   \hspace{-10pt}\includegraphics[width=\imsize]{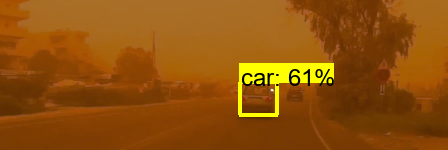}&
   \hspace{-10pt}\includegraphics[width=\imsize]{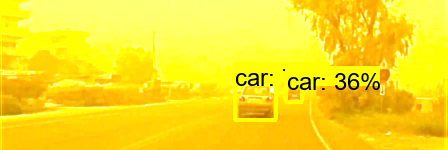}\\    
    
   (a)YOLOv3 &\hspace{-10pt}(b)GDIP &\hspace{-10pt}(c)ERUP(ours)\\
   \end{tabular}
   \caption{Visual comparison of enhanced images and detection results using (a)YOLOv3, (b)GDIP and (c)ERUP(ours). First row shows the snow results from DAWN and second and third row shows the sand results from DAWN.}
   \label{fig:conv2}
\end{figure*}

\newpage
\def\imsize{0.22\hsize}
\def\imsizeS{0.22\hsize}

\begin{figure*}[t]
   \centering
   \begin{tabular}{cccc}
   \includegraphics[width=\imsize]{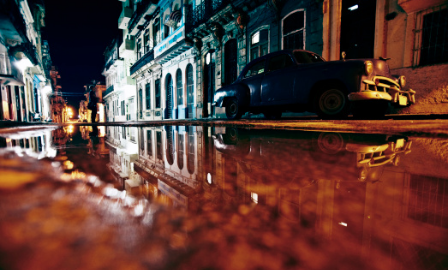}&
   \hspace{-10pt}\includegraphics[width=\imsize]{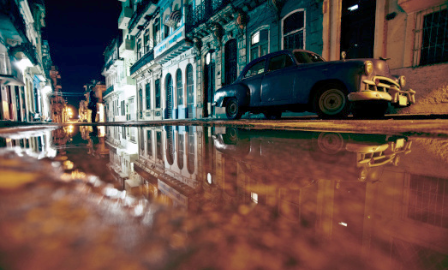}&
   \hspace{-10pt}\includegraphics[width=\imsize]{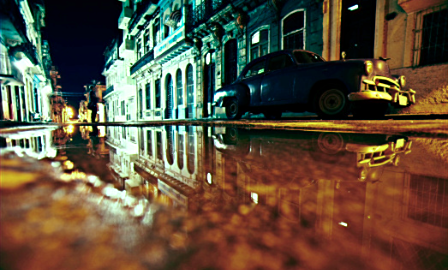}&
   \hspace{-10pt}\includegraphics[width=\imsize]{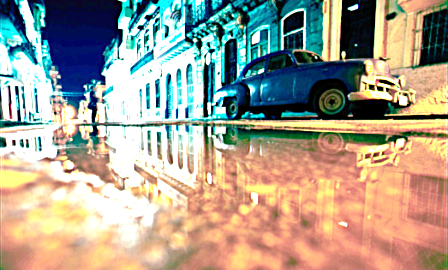}   
   \\ \vspace{-15pt} \\
   &
   \hspace{-10pt}\includegraphics[width=\imsize]{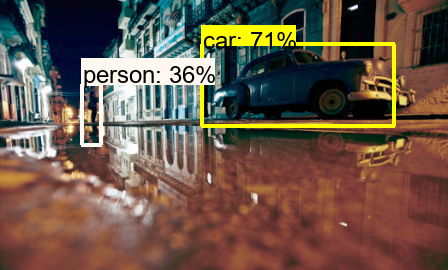}&
   \hspace{-10pt}\includegraphics[width=\imsize]{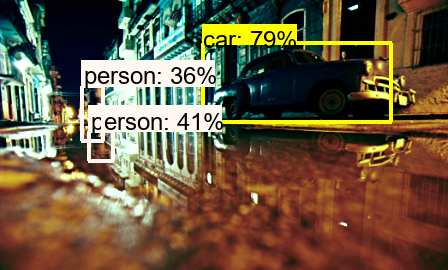}&
   \hspace{-10pt}\includegraphics[width=\imsize]{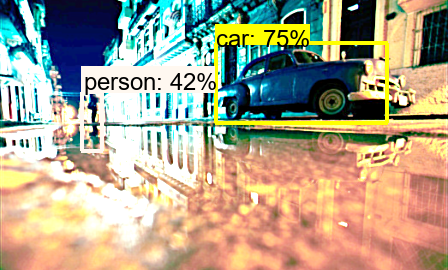}   
   \\   
   \includegraphics[width=\imsizeS]{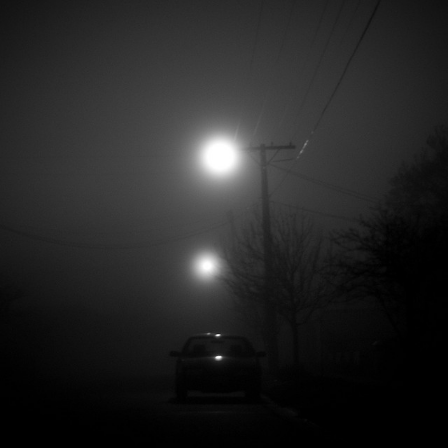}&
   \hspace{-10pt}\includegraphics[width=\imsizeS]{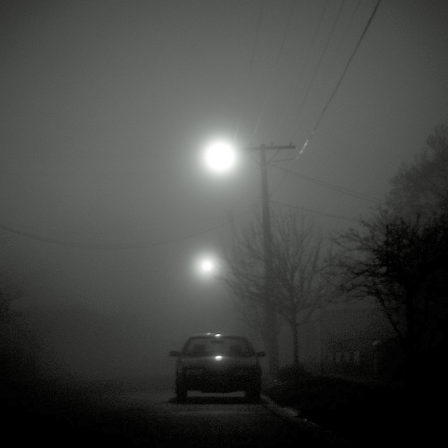}&
   \hspace{-10pt}\includegraphics[width=\imsizeS]{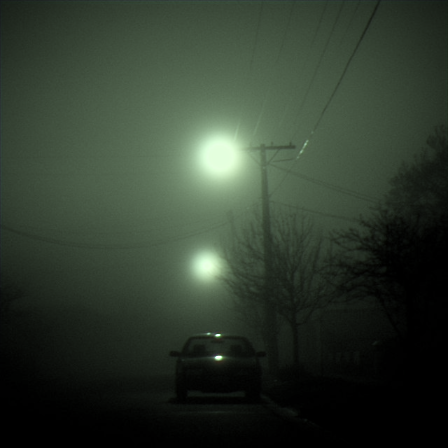}&
   \hspace{-10pt}\includegraphics[width=\imsizeS]{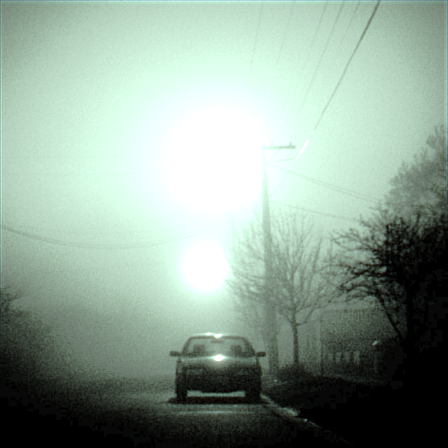}   
   \\ \vspace{-13pt} \\
   &
   \hspace{-10pt}\includegraphics[width=\imsizeS]{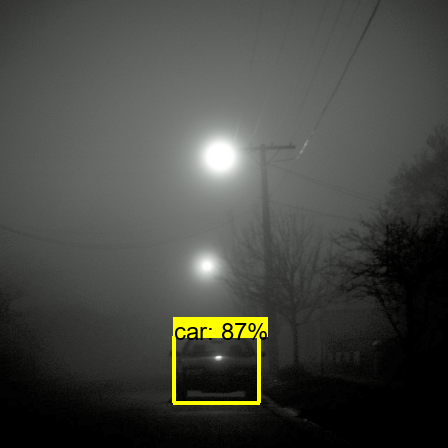}&
   \hspace{-10pt}\includegraphics[width=\imsizeS]{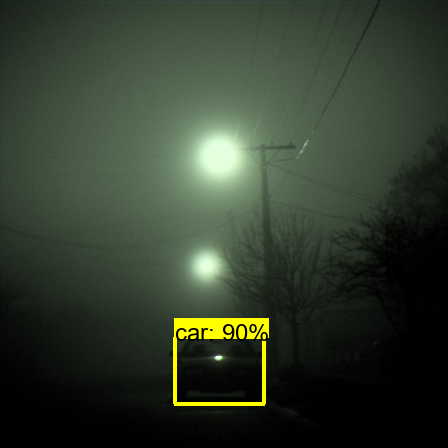}&
   \hspace{-10pt}\includegraphics[width=\imsizeS]{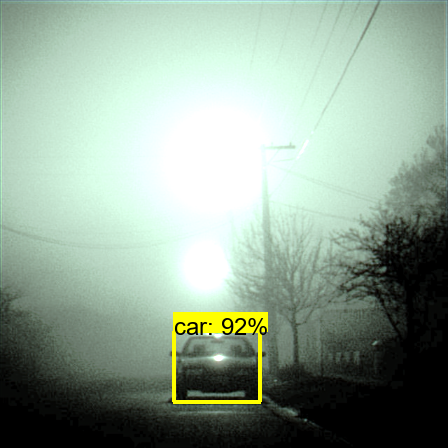}   
   \\  
   \includegraphics[width=\imsize]{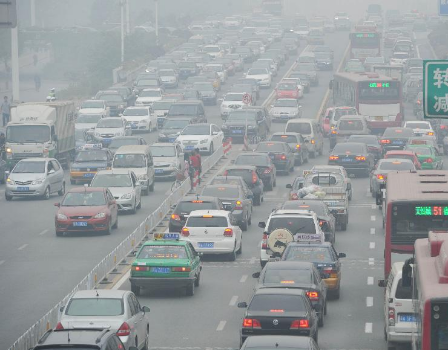}&
   \hspace{-10pt}\includegraphics[width=\imsize]{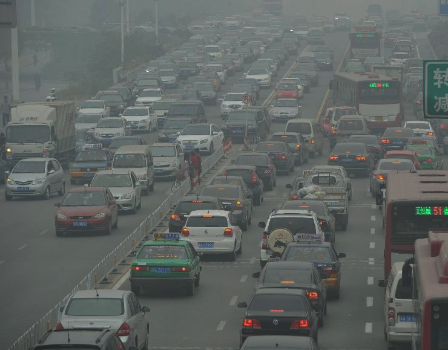}&
   \hspace{-10pt}\includegraphics[width=\imsize]{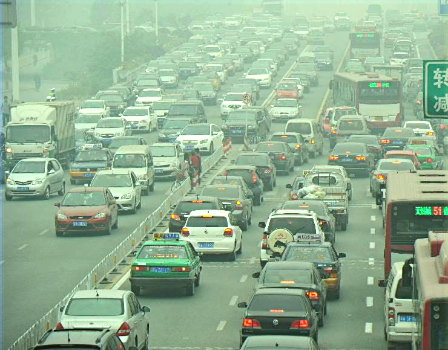}&
   \hspace{-10pt}\includegraphics[width=\imsize]{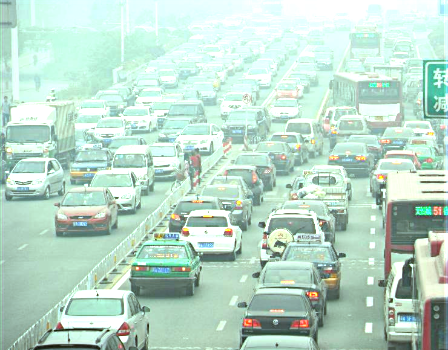}   
   \\ \vspace{-15pt} \\
   &
   \hspace{-10pt}\includegraphics[width=\imsize]{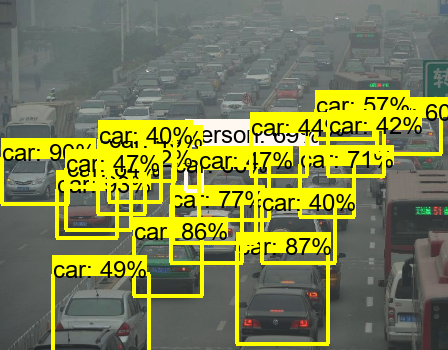}&
   \hspace{-10pt}\includegraphics[width=\imsize]{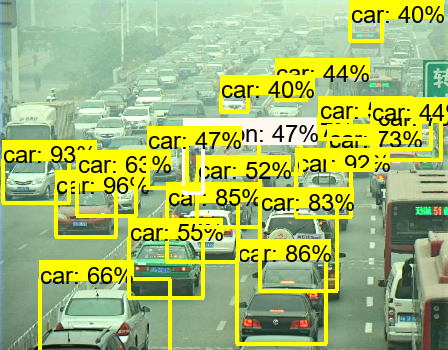}&
   \hspace{-10pt}\includegraphics[width=\imsize]{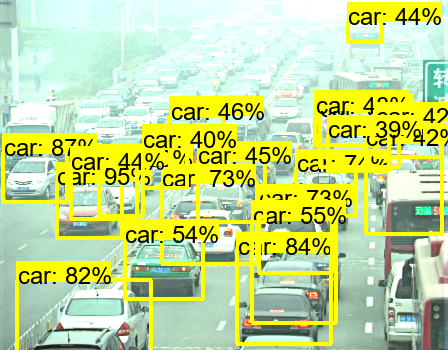}   
   \\     
   (a)Original &\hspace{-10pt}(b)BPW &\hspace{-10pt}(c)KBL&\hspace{-10pt}(d)BPW+KBL\\
   \end{tabular}
   \caption{Visual comparison of (a)Original input images, enhanced images and detection results using (b)BPW, (c)KBL and (d)BPW+KBL(proposed). First and second row shows the low-light results from ExDark, third row shows the fog results from RTTS.}
   \label{fig:abl}
\end{figure*}

\end{document}